\documentclass[acmtog, screen]{acmart}

\AtBeginDocument{%
  \providecommand\BibTeX{{%
    \normalfont B\kern-0.5em{\scshape i\kern-0.25em b}\kern-0.8em\TeX}}}

\usepackage{amsmath}
\usepackage{latexsym,amssymb,epsf}
\usepackage{mathtools}

\copyrightyear{2020}
\acmYear{2020}
\setcopyright{acmlicensed}\acmConference[ESEM '20]{ESEM '20: ACM / IEEE International Symposium on Empirical Software Engineering and Measurement (ESEM)}{October 8--9, 2020}{Bari, Italy}
\acmBooktitle{ESEM '20: ACM / IEEE International Symposium on Empirical Software Engineering and Measurement (ESEM) (ESEM '20), October 8--9, 2020, Bari, Italy}
\acmPrice{15.00}
\acmDOI{10.1145/3382494.3410675}
\acmISBN{978-1-4503-7580-1/20/10}
\begin{document}

\title{funcGNN: A Graph Neural Network Approach \\
to Program Similarity}

\author{Aravind Nair}
\affiliation{%
  \institution{KTH Royal Institute of Technology}
  \city{Stockholm}
  \state{Sweden}
  }
\email{aanair@kth.se}

\author{Avijit Roy}
\affiliation{%
  \institution{KTH Royal Institute of Technology}
  \city{Stockholm}
  \state{Sweden}
  }
\email{avijit@kth.se}

\author{Karl Meinke}
\affiliation{%
  \institution{KTH Royal Institute of Technology}
  \city{Stockholm}
  \state{Sweden}
  }
\email{karlm@kth.se}


\begin{abstract}

\textbf{Abstract:} Program similarity is a fundamental concept, central to the solution of software engineering tasks such as software plagiarism, clone identification, code refactoring and code search. Accurate similarity estimation between programs requires an in-depth understanding of their structure, semantics and flow. A control flow graph \textit{(CFG)}, is a graphical representation of a program which captures its logical control flow and hence its semantics. A common approach is to estimate program similarity by analysing CFGs using graph similarity measures, e.g. graph edit distance \textit{(GED)}. However, graph edit distance  is an NP-hard problem and computationally expensive, making the application of graph similarity techniques to complex software programs impractical. This study intends to examine the effectiveness of graph neural networks to estimate program similarity, by analysing the associated control flow graphs. We introduce \textit{funcGNN}\footnote{The code is publicly available at https://github.com/aravi11/funcGNN }, which is a  graph neural network trained on labeled CFG pairs to predict the GED between unseen program pairs by utilizing an effective embedding vector. To our knowledge, this is the first time graph neural networks have been applied on labeled CFGs for estimating the similarity between high-level language programs. We demonstrate the effectiveness of \textit{funcGNN} to estimate the GED between programs and our experimental analysis demonstrates how it achieves a lower error rate (1.94 $\times 10^{-3}$), with faster (23 times faster than the quickest traditional GED approximation method) and better scalability compared with state of the art methods. \textit{funcGNN} posses the inductive learning ability to infer program structure and generalise to unseen programs. The graph embedding of a program proposed by our methodology could be applied to several related software engineering problems (such as code plagiarism and clone identification) thus opening multiple research directions. 
\end{abstract}



\maketitle
\section{Introduction}
Finding the similarity between two objects plays an important role in many computational tasks such as recommendation and marketing. For example, well-known search engines (Google, Bing), e-commerce sites (Amazon, e-Bay) and online media service providers (YouTube, Netflix), all invest heavily in similarity measures to recommend the next best product for their users. Another area where similarity measures are helpful is in software engineering. Program similarity or code similarity is a fundamental theoretical concept, central to the effective solution of software engineering tasks such as software plagiarism, clone identification, code refactoring and code search. The basic idea of a program similarity metric is to quantitatively measure how one program is syntactically similar to another program. Program similarity is related to program equivalence, where the latter concept usually refers to semantic similarity. However, program equivalence is a more challenging task as a minor change in code structure can drastically change its logic making it structurally similar but semantically very different. Since semantic equivalence of programs is undecidable, program similarity represents a simpler but more tractable syntactic approximation. 

One of the most widely used techniques to analyse programs is by transforming them into graphs \cite{allamanis2017learning}.  A graph is a mathematical structure used to represent the relationship and connection between objects termed nodes. Graph structures are ubiquitous in real-life and can be found in almost every domain. The most frequently used graph representations in the field of program analysis are control graphs \textit{(CG)}, abstract syntax trees \textit{(AST)}, control flow graphs \textit{(CFG)} and program dependency graphs \textit{(PDG)}. A CFG is a directed graph in which each node represents an atomic operation or statement, and each edge represents a possible transfer of control (i.e. execution order). As CFGs are capable of preserving both the logic flow and semantics of a program, we can use the CFG representation to address the problem of program similarity. We have used Soot\footnote{https://github.com/Sable/soot} as an open source bytecode manipulation and optimization framework to generate CFGs for Java programs. Soot processes the bytecode of a Java program and converts it into an intermediate representation \textit{Jimple} \cite{vallee1998jimple}.  \textit{Jimple} breaks down a statement to a 3-address instruction set to provide a detailed atomic operation model of the program. These atomic operations label the nodes of the CFG. Figure 1 shows the transformation of a simple Java function (sum of all elements in an array) to its corresponding CFG. 
\begin{figure}
    \centering
    \includegraphics[scale=0.45]{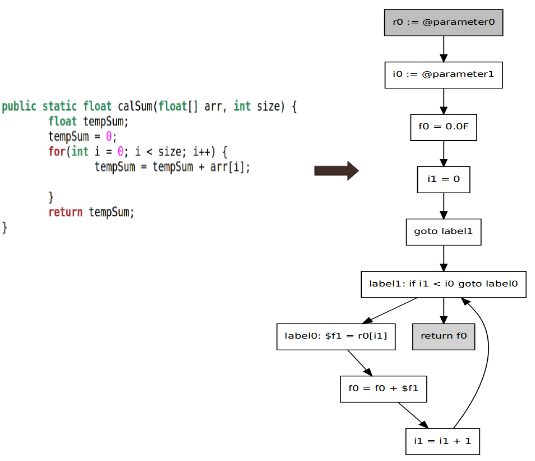}
    \caption{Java function to calculate sum of all elements in an array and its corresponding CFG}
    \label{fig:Fig 1}
\end{figure}

Though graphs provide good program models, graph similarity is not an easy task. Graph similarity metrics like Graph Edit Distance \textit{(GED)} \cite{bunke1997relation} and Maximum Common Subgraph \textit{(MCS)} \cite{bunke1998graph} are known to be NP-hard problems, hence computationally expensive. Using the GED metric for large graphs is impractical, as one cannot compute the similarity score between graphs of more than 16 nodes within a reasonable time \cite{blumenthal2018exact}.
Over the past ten years, graph neural networks (\textit{GNN}) have emerged in machine learning (ML) as a successful class of neural network models. GNNs are capable of supervised learning of graph representations to efficiently implement many relations and functions on graphs. In one approach, a GNN generates a meaningful vector embedding for each node in a graph using a recursive neighborhood aggregation algorithm \cite{scarselli2008graph}. This embedding approximates the semantics of the corresponding nodes and can be used for supervised machine learning tasks like node classification, graph classification, graph similarity etc.    

In this research, we aim to investigate whether program similarity could be estimated by analysing labeled control flow graphs using a graph neural network. We introduce \textit{funcGNN},  a graph neural network trained to predict the GED between program pairs. In our approach, we use an amalgamation of two graph embedding techniques. The first technique is a \textit{top-down} approach which creates an embedding for the whole graph by transforming into a meaningful fixed size vector. To make sure that this embedding vector captures the semantic information of semantically significant nodes we incorporate an attention mechanism into the neural network architecture. 
The second technique, on the other hand, uses a \textit{bottom-up} approach. This focuses on node level comparison of graph pairs and captures the atomic program operation similarities. Using these atomic node-level similarities, we compute a histogram feature representation which gives an inferred probability distribution for node similarity. The two vector embeddings generated by these two techniques are concatenated, and the resulting embedding is the input vector for multiple fully connected neural network layers to predict the similarity score between a pair of graphs. The similarity score thus predicted is the normalized GED score between the graph pair. 

To our knowledge, this is the first time graph neural networks have been applied on labeled CFGs of high level languages for program similarity. We evaluate the effectiveness of our proposed methodology on functions from open source Java code. 

The main contributions of this research can be summarized as follows:
\begin{itemize}
    \item We address the problem of program similarity and propose \textit{funcGNN}, a novel graph neural network capable of predicting the similarity between program pairs by analysing their labeled control flow graphs.
    \item We use two graph embedding techniques with an attention mechanism and histogram feature representation to calculate the overall program similarity. 
    \item We empirically demonstrate how \textit{funcGNN}: (i) generalises well to unseen graph program pairs, (ii) achieves low error rates and (iii) significantly reduces computation times compared to the state of the art GED approximation methods. Thus our solution has better scaling properties.
\end{itemize}

 The rest of this paper is organised as follows. We briefly explain the theoretical prerequisites for our work in Section 2. This includes a brief introduction to control flow graphs, graph edit distance and graph neural networks. Section 3 describes in detail \textit{funcGNN}, our proposed ML architecture along with its sub-modules and their algorithms. In Section 4, we describe our evaluation dataset for \textit{funcGNN}, and the hyper parameters of \textit{funcGNN}. We present the experimental results obtained on our dataset. We discuss the limitations and threats to validity of our study in Section 5. Section 6 provides a survey of relevant literature. We conclude the paper in Section 7 with a summary of our results and a discussion of future directions for research.  
\section{Background}

\subsection{Control Flow Graphs}
To estimate the similarity between two programs, we will compare their control flow graphs\footnote{https://en.wikipedia.org/wiki/Control-flow\_graph}. A \textit{CFG} is a graph representation which specifies the logic and control flow of a program. A control flow graph can be represented by a directed graph $G =(V,E \subseteq V \times V)$, where $V$  denotes the set of nodes and E the set of edges (node pairs) connecting them. Each node $v_i \in V$, is labelled by an atomic program statement. A pair of nodes $v_{i}$, $v_{j}$ is connected by an edge $(v_{i}, v_{j} ) \in E$, when this reflects the direct execution order: $v_{i}$ is immediately followed by $v_{j}$. Thus a CFG provides a graph representation of both program syntax and semantics\footnote{A CFG can be compared with an abstract syntax tree which represents only program syntax.}. 

\subsection{Graph Edit Distance and its Approximations}
Though graphs are ubiquitous in almost every field of computer science, finding similar graphs is a challenging task. There are well defined graph similarity metrics like graph edit distance (GED) \cite{bunke1997relation} and maximum common subgraph (MCS) \cite{bunke1998graph}. In this research we use GED as the similarity metric between two graphs. Graph edit distance is analogous to the edit distance concept used for string matching \cite{ristad1998learning, cohen2003comparison}. The GED of two graphs can be defined as the number of operations required to transform one graph to another. Formally, given two graphs $G_1$ and $G_2$, then the GED between them can be defined by,
\begin{equation}
GED_{(G_1, G_2)} = min_{(x_i,.., x_k) \in P(G_1, G_2)} \sum_{i=0}^{k} c(x_i)
\end{equation}
where, $P(G_1, G_2)$ denotes the set of all possible edit paths (sequences of atomic edit operations) that transform $G_1$ to $G_2$. Here, $c(x) \geq 0$ denotes the cost of each graph edit operation x, which includes deletion, insertion and substitution of nodes and edges. A classical approach to calculate the GED between two graphs is described in \cite{neuhaus2006fast}, in which they identify the minimal edit path using the A* algorithm. However this approach has exponential time complexity and several studies have been made to reduce its execution time \cite{bougleux2017graph, riesen2009approximate, zeng2009comparing, fankhauser2011speeding, neuhaus2007quadratic, riesen2014improving}.  

In this study, we estimate the approximate GED between graph pairs by using the Quadratic Assignment Problem (\textit{QAP}) approximation as proposed in \cite{bougleux2017graph}. The inspiration to substitute approximate values in place of the exact GED for large graphs comes from the ICPR2016\footnote{https://gdc2016.greyc.fr/} contest. Once the approximate graph edit distance GED between a graph pair is calculated, we normalize it in a range between 0 and 1. This normalized score acts as the similarity score or metric for that graph pair. A detailed explanation of the normalization function is provided in Section \ref{section:data_explanation}

\subsection{Graph Neural Networks}
The graph neural network (GNN) model was designed as an extension of the recurrent neural network model. The goal 
was to efficiently learn relations and functions on graphs \cite{gori2005new, scarselli2008graph}. The concept of GNN is based on the idea that each node in a graph $G = (V,E)$ can be characterised in terms of: (i) its own features, (ii) the relations it has with its locally neighbouring nodes, and (iii) the features of its local neighbours. To represent each node $v \in V$, a GNN uses an s-dimensional state embedding vector $h_v \in \mathbb{R}^s$ which consists of information about the node and its neighbours. The state embedding $h_v$ is learned via a parametric non-linear local transition function \textbf{$f_{lt}$}, which is uniformly defined parametrically across all nodes. This local transition function \textbf{$f_{lt}$} captures the above three characteristics of a node and its local neighbors. The state embedding $h_v$ can be mathematically defined as,
\begin{equation}
     h_v = f_{lt} (x_v, x_{co[v]}, x_{ne[v]}, h_{ne[v]})  \label{eq:2}
\end{equation}
where, $x_v$ represents the features of node v, $x_{co[v]}$ represents the features of the edges of node v,  $x_{ne[v]}$ represents the features of neighboring nodes of v and $h_{ne[v]}$ represents the states of the neighboring nodes of v. The state embedding $h_v$ along with the node feature $x_v$ is used to learn the final representation $o_v$ of the node v, using a parametric non-linear local output function $f_{lo}$. This final representation $o_v$ is defined as follows: 
\begin{equation}
      o_v = f_{lo} (h_v, x_v) \label{eq:3}
\end{equation}

Let $H, O, X$ and $X_n$ be the tensor vectors generated by stacking the state, output, all features and individual node features $v \in V$ in the graph $G$. Then equations \ref{eq:2} \& \ref{eq:3} can be written as: 
\begin{equation}
      H = F_{GT} (H, X) \label{eq:4}
\end{equation}
\begin{equation}
      O = F_{GO} (H, X_n) \label{eq:5}
\end{equation}
where, $F_{GT}$ representing the global transition function, and $F_{GO}$ representing the global output function are the stacked versions of $f_{lt}$ and $f_{lo}$
respectively. A unique solution to equations \ref{eq:4} \& \ref{eq:5} can be found by using Banach's Fixed Point Theorem \cite{khamsi2011introduction}. According to this theorem, a unique solution can be calculated as a fixed point of the operators $F_{GT}$ and $F_{GO}$ provided that these are \textit{contraction maps}
with respect to the state. The contraction condition 
means that there exists some $0 \leq \mu < 1$ such that,
\begin{equation}
    ||F_{GT}(H,X) - F_{GT}(I,X)|| \leq  \mu||H-I|| \label{eq:6}
\end{equation}
for any H,I, where $||.||$ denotes the vector norm on states. Besides guaranteeing a unique solution, Banach's Theorem actually gives an iterative scheme for approximating the fixed-point: 
\begin{equation}
    H^{t+1} = F_{GT}(H^{t}, X)  \label{eq:7}
\end{equation}
where, $H^{t}$ denotes the $t^{th}$ iteration of H. For any initial value of $H(0)$, the dynamical system of equation \ref{eq:7} converges exponentially fast to a fixed point solution of equation \ref{eq:4}. Thus, $H(t)$ denotes the state that is updated by the global transition function $F_{GT}$. Equation \ref{eq:7} could then be rewritten as, 
\begin{gather} \label{eq:8}
   h_v^{t+1} = f_{lt} (x_v, x_{co[v]}, x_{ne[v]}, h_{ne[v]}^t   )\\
  o_v^t = f_{lo} (h_v^t, x_v)   
\end{gather}

By keeping the target information $t_v$ for each node $v$, GNNs learn the parameters of $f_{lt}$ and $f_{lo}$ by minimising the loss between the targeted value $t_v$ and the output value $o_v$. This loss function can be represented as, 
\begin{equation}
    loss = \sum_{i=1}^{N} (t_v - o_v)^2 
\end{equation}
where $N$ is the number of nodes in the graph $G$. The learning task is carried out iteratively based on a gradient-descent strategy until time $T$, where the fixed point solution of equation \ref{eq:4} is achieved i.e. $H(T) \approx H$.
 
\section{Proposed Methodology}

\begin{figure*}
    \centering
    \includegraphics[scale=0.3]{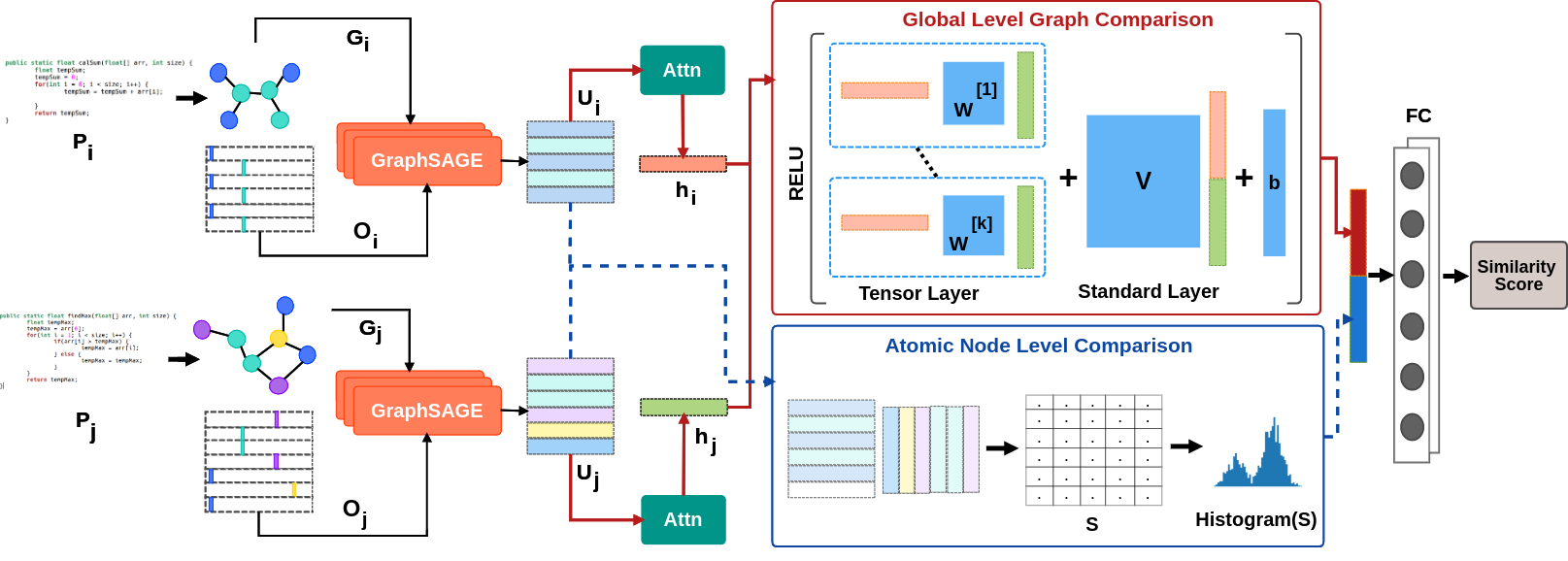}
    \caption{Overall architecture and workflow of \textit{funcGNN}}
    \label{fig:Fig 2}
\end{figure*}

In this section, we first describe the \textit{funcGNN} framework to learn the approximate GED between program pairs. Given two graphs $G_1$ and $G_2$, \textit{funcGNN} creates a fixed size vector embedding for each graph, and learns a model of the similarity function, which could map these input embeddings to a single real-valued similarity score. To generate the embeddings, we use a combination of the whole graph embedding (a top-down approach) and atomic node level comparison representation (a bottom-up approach). The embeddings generated by these two methods are concatenated and fed to multiple fully connected neural network layers to calculate the similarity score. Figure \ref{fig:Fig 2} depicts the overall architecture and workflow of \textit{funcGNN}. A detailed explanation of this architecture is given in the sequel.    
 \subsection{Top-down Approach - Overall Graph Embedding}
 In the top-down approach, we aim to efficiently create a global embedding for the entire CFG capturing its global structure and control flow. This global embedding of a CFG is then used to predict program similarity. This global approach involves the following stages: 
\subsubsection{\textbf{Inductive Node Embedding:}}\label{node_embedding}
 For generating the embedding of each node $v$ in the CFG, we use the GraphSAGE method as proposed in \cite{hamilton2017inductive}. GraphSAGE is an inductive node embedding method which generalizes to unseen nodes, which is one of the major problems when analysing a new program not previously seen in the training data. The GraphSAGE methodology is different from the original GNN approach \cite{scarselli2008graph} as it defines the neighborhood of a node by aggregating the features from a sampled subset of its entire neighborhood. This could be represented as,
 \begin{equation}
     h^t_{N_v} = AGG_t\,({h^{t-1}_u), \,\forall u \in N_v})  \label{eq:11}
\end{equation} 
\begin{equation}
      h^t_v = \sigma\,(W_t \cdot [\,h^{t-1}_v \,||\, h^t_{N_v}\,]) \label{eq:12}
 \end{equation}
 where $N_v$ represents the neighborhood set of node $v$, $AGG_t$ represents the aggregate function, $W$ represents the weight matrix, $||$ represents the concatenation operation, $\sigma$ represents the non-linear activation function and $h^t_v$ denotes the state embedding of node $v$ at time t. We use the \textit{mean aggregator} as the aggregator function $AGG_t$ which is an approximation of the convolution operation in the GCN framework proposed in \cite{kipf2016semi}. The \textit{mean aggregator} function is a variant of the skip connection \cite{he2016identity} and does not perform the concatenation operation as in equation \ref{eq:12}. Thus equation \ref{eq:12} could be rewritten as, 
 \begin{equation}
     h^t_v = \sigma\, (W \cdot \textsc{mean}\,(\{h^{t-1}_v\} \cup \{\,h^{t-1}_u, \,\forall u \in N_v\}))
 \end{equation}
 In our approach, the nodes are initially represented by the one-hot embedding scheme $O \in \mathbb{R}^{N \times D}$ where $N$ is the number of nodes in the graph and $D$ is the dimension size. This one-hot embedding is then passed through multiple layers of GraphSAGE to get their node representation $U \in \mathbb{R}^{N \times D}$. We set the GraphSAGE layers to 3 as GNNs have an over-smoothing issue with deep architectures \cite{li2018deeper} and we use \textit{relu} as our activation function. 

 \subsubsection{\textbf{Attention Based Graph Embedding}} Once we obtain the representations of each node $U$, the next task is to combine them effectively to generate an overall embedding $h$ for the whole graph. Instead of simply averaging the embedding of all the nodes in the graph we use an attention mechanism to provide more significance to certain nodes based on a similarity metric. This approach makes sure that nodes with more structural significance have more influence on the overall graph embedding compared to the other nodes. From a program analysis point of view, the idea is to provide more weight to nodes labelled with mathematical operations than nodes labelled with variable initialization or assignment.  
 
To achieve this, we first compute a context vector embedding $c$ by averaging all the node embeddings $U$ and transforming them through a non-linear \textit{relu} activation function. This graph context vector $c$ can be represented as, 
\begin{equation}
    c = \textit{relu}\, (\,(\frac{1}{|V|}\sum_{v=1}^{|V|} U_v) \cdot W )
\end{equation}
where $|V|$ is the number of nodes in the graph and $W$ is a weight matrix of dimension $W \in \mathbb{R}^{D \times D}$.
By learning the weight matrix $W$, the context vector $c$ provides a naive summary of the structural attributes of the entire graph. To calculate the attention weight $a_v$ of each node $v$ we take the inner product of each node embedding $U_v$ with the context vector $c$. The idea behind this approach is that majority of the node operations in a program will have mathematical operations rather than variable initialization or assignment operations. Hence the node embeddings for mathematical operations will have more impact on the context vector $c$, and node embeddings most similar to the context vector $c$ will attain higher attention weight. Once we calculate the attention weight $a_v$ for each node, we calculate the overall graph embedding $h \in \mathbb{R}^D$ by, 
\begin{gather}
    a_v = \textit{relu}\,(U^T_v \cdot c) \\
    h = \sum^{i=1}_{N} (a_v \cdot U_v)
\end{gather}

\subsubsection{\textbf{NTN based Graph Pair Comparison:}}
Now that we have obtained the overall embedding $h$ for each graph, the challenge is to efficiently compare two graph embeddings to estimate their similarity. For this we use the \textit{Neural Tensor Network (NTN)} as proposed in \cite{socher2013reasoning}. The advantage of the \textit{neural tensor network (NTN)} over the traditional linear layer approach is its ability to efficiently compare two embedding vectors across multiple dimensions. Given two embedding vectors $h_i$ and $h_j$, NTN makes use of a bilinear tensor, which computes the relationship between the two embeddings using the following function, 
\begin{equation}
    NTN_{(h_i, h_j)} = \sigma \,(h^T_i \cdot W^{[1:k]} \cdot h_j + V_R \begin{bmatrix} h_i \\ h_j \end{bmatrix} + b )
\end{equation}
where $\sigma$ is a non-linear activation function, $W^{[1:k]} \in \mathbb{R}^{d \times d \times k}$ is a tensor vector with $k$ slices, $V \in \mathbb{R}^{k \times 2d}$ is the weight matrix of a standard neural network and $b$ is the bias. The bilinear tensor product $h^T_i \cdot W^{[1:k]} \cdot h_j$ computes a representation vector $S \in \mathbb{R}^k$ where each slice of the tensor learns a distinct pattern of similarity between the input embeddings.
 
\subsubsection{\textbf{Similarity Score Estimation}}
The final task is to generate the similarity score $\hat y_{i,j}$, by passing $NTN_{(h_i, h_j)}$ through multiple layers of fully connected neural networks. A fully connected layer is a non-linear neural network which maps the input of dimension $m$ to a desired output dimension $n$ using multilayered weighted neuron multiplications.  From the final layer of the fully connected neural network, one score is calculated which represents the $\hat y_{i,j}$. In the training phase, we compare the generated  $\hat y_{i,j}$ with the actual ground truth $y_{i,j}$ and minimize the \textit{mean square error} (MSE) loss using a gradient-descent strategy. This can be represented as, 
\begin{equation}
     loss_{mse} = \frac{1}{D} \sum_{(i,j) \in D} ( \hat y_{i,j} - y_{i,j})^2  \label{eq:18}
\end{equation}
 where D is the number of of graph pairs in the dataset.

 \subsection{Bottom-Up Approach - Atomic Level Node Comparison}
When generating an overall embedding of a graph using the top-down approach, there is the possibility of losing some of the local node information. To overcome this, we also use a \textit{bottom-up approach}. Here, instead of comparing the entire graph embedding we match the similarity between nodes among the graph pair. The idea is to extract atomic node level similarity in an analogous way to the methodology of random walk kernels\cite{neuhaus2006random} on graphs. However, graph kernels are computationally expensive and at least of the order $O(n^3)$ \cite{vishwanathan2006fast}. To obtain the node similarities we instead take the inner product of all the pairwise node combinations of the two graphs using their embedding as discussed in Section \ref{node_embedding}. The inner product score is later transformed by passing it through a non-linear activation function. To make sure that the graphs are of the same size, we pad the shortest graph in the pair with nodes having zero embedding, i.e. an embedding with zero vector initialization, until both graphs are of the same size. The result of the inner product is thus a similarity matrix, $S \in \mathbb{R}^{N \times N}$, where $N = max(N_{G_i}, N_{G_j})$ and $N_{G_i}$ denotes the number of nodes in graph $G_i$. 

To efficiently utilise the atomic level node pair similarity we transform the similarity matrix $S$ into a histogram feature vector $H(S) \in \mathbb{R}^b$, where b defines the number of bins. Histograms provide the probability distribution of the node similarities in $S$ and are invariant to the node ordering. This is the same issue which causes the graph isomorphism problem to have high computational complexity. The histogram feature vector $H(S)$ thus obtained is passed to the fully connected layers after normalization and concatenated with $   NTN_{(h_i, h_j)}$, to calculate the final similarity score.

\section{Experimental Design and Results of Training and Evaluation}

In this section, we describe the training and evaluation of  \textit{funcGNN} on a labeled CFG dataset derived from open source Java programs. We describe our experimental setup and present the results obtained from this empirical study. 

\subsection{The CFG Dataset}
\label{section:data_explanation}
\begin{figure*}
    \centering
    \includegraphics[scale=0.18]{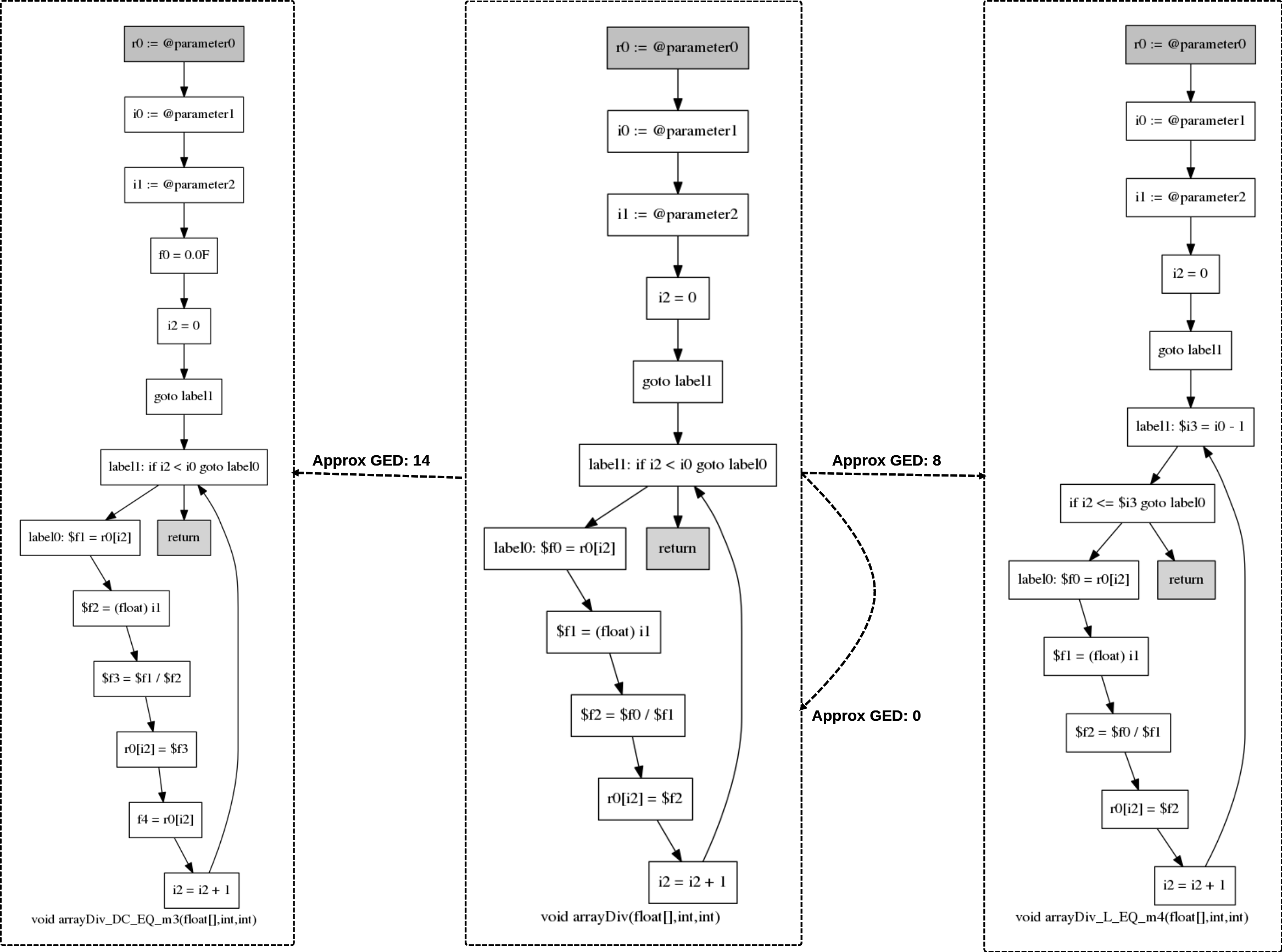}
    \caption{Approximate GED values between array division program and two of its mutants}
    \label{fig:Fig 3}
\end{figure*}

\begin{figure}
    \centering
    \includegraphics[scale=0.52]{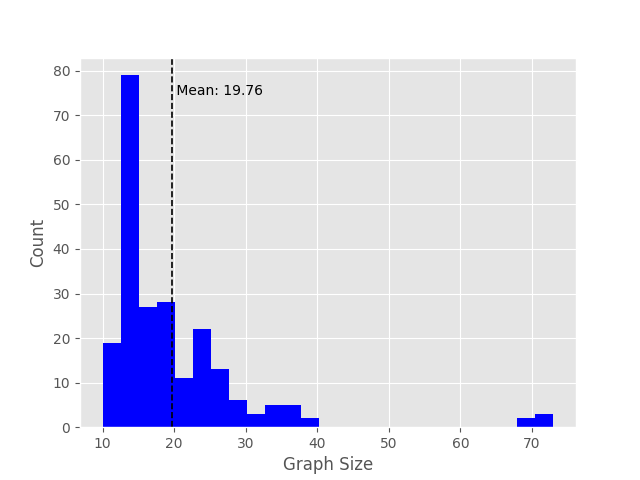}
    \caption{Distribution of graph sizes in the dataset}
    \label{fig:Fig 4}
\end{figure}

\begin{figure*}
    \centering
    \includegraphics[scale=0.35]{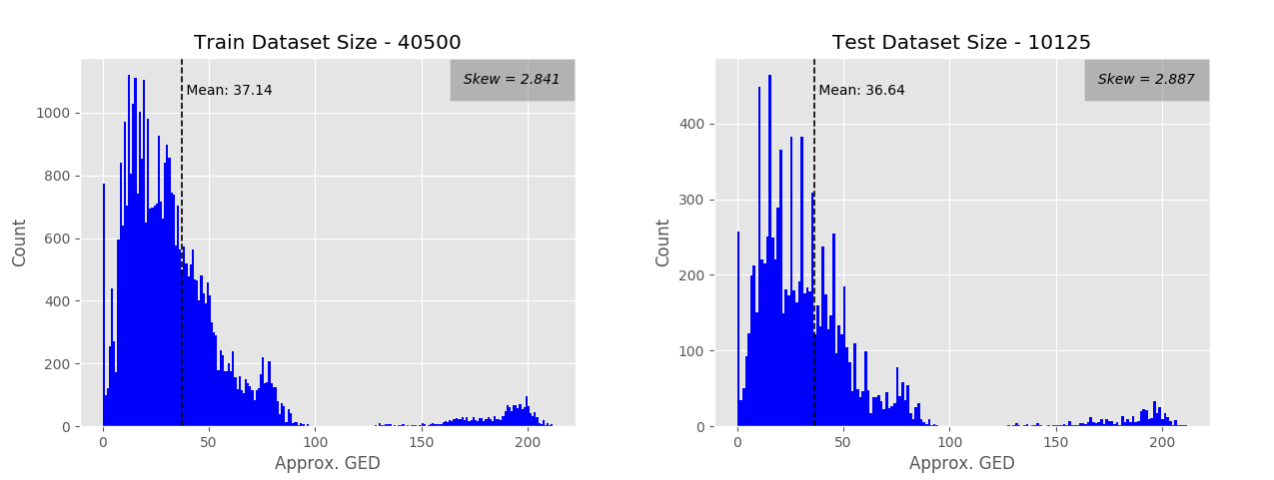}
    \caption{Distribution of Approx GED values in the training and testing sets}
    \label{fig:Fig 5}
\end{figure*}

For the task of learning program similarity we collected a set of 45 open source Java functions (such as Bubble Sort). These are rich in mathematical operations. Since these programs were all distinct algorithms, their GED values were very high. To overcome this issue, we generated mutants of these programs as a data augmentation method first presented in \cite{nair2019leveraging}. A mutant is a structurally modified version of a program, usually created by injecting a fault into it \cite{geist1992estimation}, e.g. by changing an operator or relation. By restricting change to one operator, each mutant has a low GED from its original program. Thus we created a dataset having program pairs with both high and low GED values. We generated 4 mutants of each program thus extending our dataset to 225 java functions. To extract the CFGs from these programs we used the Soot framework as shown in Figure \ref{fig:Fig 1}. We constructed a graph pair dataset of 50625 program pairs in a JSON format which included the node labels, node attributes, edge list and their approximate edit distance using the QAP method. Figure \ref{fig:Fig 3} shows the approximate GED values between array division program and two of its mutants. Figure \ref{fig:Fig 4} shows the graph size distribution of our dataset. We split our dataset using an 80:20 ratio into training and testing datasets. Figure \ref{fig:Fig 5} depicts the approx GED distribution in both the training and testing sets.  
In our study we chose the approximate GED value as the ground truth value for each graph pair. We transformed these ground truth GED values to the ground truth similarity score $S_{GT}$ by first normalizing them and then passing through an exponential function to map into the interval $(0,1]$. This can be represented as, 
\begin{gather}
   norm = 2 \cdot \frac{GED(G_i, G_j)}{|G_i| + |G_j|} \\
   S_{GT} = e^{-norm}
\end{gather}


\subsection{Results }
All experiments were conducted on a single mini workstation: 2.60GHz Intel(R) Xeon(R) CPU E5-2697 v3, 56 CPU cores and 250 GB RAM. To evaluate and compare \textit{funcGNN} with state-of-the art methods, we used the following two metrics:
\begin{itemize}
    \item \textit{MSE}: We used the mean square error (MSE) to compute the loss between the predicted score and the ground truth, as defined in equation \ref{eq:18}. MSE satisfies the mathematical properties of convexity, symmetry, and differentiability, and is sensitive towards outliers in a dataset. Figure \ref{fig:Fig 6} shows the \textit{MSE} loss function curve obtained by \textit{funcGNN} for the training and testing datasets. From figure \ref{fig:Fig 6} we can infer that the proposed funcGNN model has converged deftly for both the train and test datasets. The performance and convergence behavior of the model indicate that MSE is a good approach for optimizing the funcGNN model for learning the similarities between programs. We compared the MSE error value obtained by funcGNN with other traditional graph edit distance algorithms and graph neural networks in table \ref{table:1}. Our experiments show that funcGNN surpasses the traditional methods and provides a better generalised model for finding similarity between programs. 
\begin{figure*}
    \centering
    \includegraphics[scale=0.5]{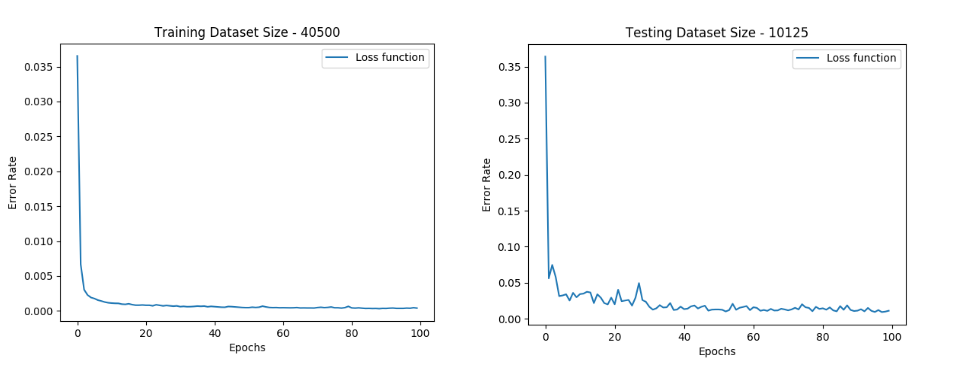}
    \caption{Loss function for both training and test sets }
    \label{fig:Fig 6}
\end{figure*}

\begingroup
\setlength{\tabcolsep}{8pt} 
\renewcommand{\arraystretch}{1.2} 
\begin{table}
\centering
\begin{tabular}{ |p{2.5cm}|p{2cm}|} 
 \hline
 
 \hline
 Method & MSE in $10^{-3}$\\
 \hline
 QAP \cite{bougleux2017graph}   & 0.0   \\
 VJ \cite{fankhauser2011speeding} & 14.41  \\
Hungarian \cite{riesen2009approximate} & 15.97 \\
  HED \cite{fischer2017improved} & 8.67 \\
 \hline
 GCN \cite{kipf2016semi} & 4.58 \\
 GraphSAGE \cite{hamilton2017inductive} & 3.61  \\
 \textbf{funcGNN}&  \textbf{1.94}  \\
 \hline
\end{tabular}
\caption{Comparison results of the mse error rate}
\label{table:1}
\end{table}
\endgroup  

    \item \textit{Time}: By time we mean the total time taken by each method to estimate the similarity score of the graph pairs for the entire dataset. Table \ref{table:2} shows the \textit{time} comparison of \textit{funcGNN} along with other metrics to predict the GED values of all the graph pairs. Since GED estimation is time consuming, we harnessed the power of parallel computing to speed up the computation. The approximate GEDs for traditional approaches were calculated  asynchronously via ProcessPoolExecutor\footnote{https://docs.python.org/3/library/concurrent.futures.html} \cite{hunt2019futures} using a pool of 45 concurrent processes. Our results show that \textit{funcGNN} even on serial execution provides faster results than the parallel execution (45 processes) of all the approximation methods.

\begingroup
\setlength{\tabcolsep}{8pt} 
\renewcommand{\arraystretch}{1.2} 
\begin{table}
\centering
\begin{tabular}{ |p{2.5cm}|p{1.3cm}|p{1.3cm}|} 
 \hline
 \multicolumn{3}{|c|}{Time taken in seconds }\\
 \hline
 Method & Time & \#Parallel Process\\
 \hline
  QAP \cite{bougleux2017graph}  & 9044.72 & 1  \\ 
  \hline
 QAP \cite{bougleux2017graph}  & 405.86 & 45  \\ 
 VJ \cite{fankhauser2011speeding}&   2513.73 & 45 \\ 
 Hungarian \cite{riesen2009approximate} & 2546.54& 45\\
 HED \cite{fischer2017improved}    & 9880.18 & 45\\ 
  \hline
  \textbf{GCN} \cite{kipf2016semi} & \textbf{378.96} & \textbf{1}\\
  GraphSAGE \cite{hamilton2017inductive} & 379.24 & 1 \\
 funcGNN &  379.81 & 1 \\
 \hline
\end{tabular}
\caption{GED prediction runtime comparison}
\label{table:2}
\end{table}

\endgroup  
\end{itemize}

\subsection{Case Studies }
We demonstrate three case studies of the predictions made by the proposed funcGNN model. All the case study examples are taken from the test dataset and hence unseen by the trained model. In the case study examples, we explain the performance of funcGNN when applied on program pairs having high and low similarities. We also provide an example where funcGNN has a high error value leading to a wrong prediction. The results of all the case studies are consolidated and presented in Table \ref{table:3}

\subsubsection{Case Study 1 : Program pairs with high similarity}
We wanted to analyse the ability of funcGNN to learn program pairs with high similarity score. The case were two programs will have the highest similarity value will be when they are identical. Hence we randomly chose a program pair (\textit{elementwiseMax\_DC\_EQ\_m3}, \textit{elementwiseMax\_DC\_EQ\_m3}) from our test dataset in which both of the programs are the same. \textit{elementwiseMax\_DC\_EQ\_m3} is an equivalent mutant version of \textit{elementwiseMax} program generated using the methodology described in \cite{nair2019leveraging} and outputs the elementwise maximum of two arrays. Figure \ref{fig:Fig 7} depicts the control flow graph of elementwiseMax\_DC\_EQ\_m3 function. Since both the programs in the pair are the same, the ground truth similarity score between them is 1. The proposed funcGNN model predicted the similarity value for this pair as 0.9723 with an error of 0.026

\subsubsection{Case Study 2 : Program pairs with low similarity}
\begin{figure*}
    \centering
    \includegraphics[scale=0.6]{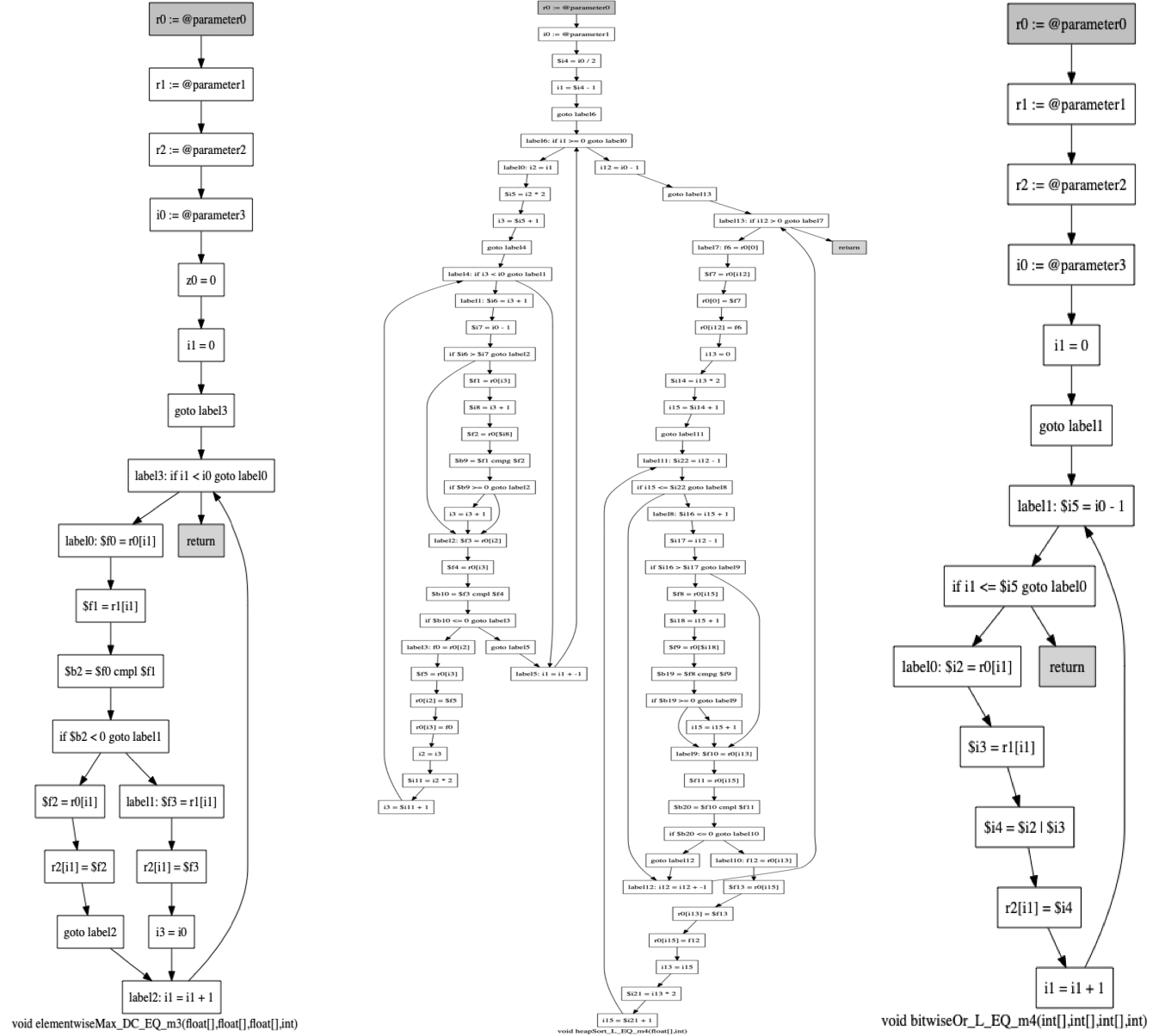}
    \caption{Control flow graphs of elementwiseMax\_DC\_EQ\_m3, heapSort\_L\_EQ\_m4 and bitwiseOr\_L\_EQ\_m4 functions}
    \label{fig:Fig 7}
\end{figure*}

We chose a program pair in the test dataset which had low similarity score. We chose the pair ($\textit{heapSort\_L\_EQ\_m4}$, $\textit{bitwiseOr\_L\_EQ\_m4}$) which had a ground truth similarity score of 0.0108. Figure \ref{fig:Fig 7} shows the control flow graphs of heapSort\_L\_EQ\_m4 and bitwiseOr\_L\_EQ\_m4 functions respectively. The proposed funcGNN model predicted the similarity value for this pair as 0.0035 with an error of 0.0073

\subsubsection{Case Study 3 : Program pair which received high error rate}
\begin{figure}
    \centering
    \includegraphics[scale=0.65]{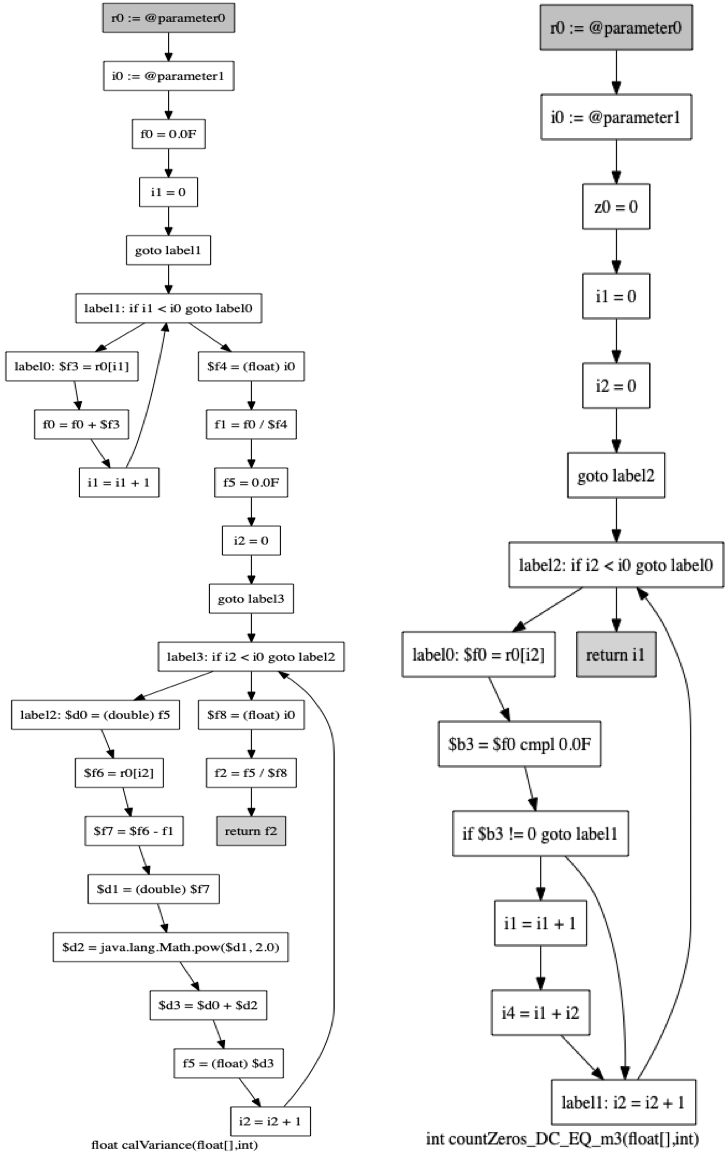}
    \caption{Control flow graphs of calVariance and countZeros\_DC\_EQ\_m3 functions}
    \label{fig:Fig 9}
\end{figure}

Here we demonstrate a program pair in the test dataset which received highest error score between their ground truth and prediction score. The pair (\textit{calVariance}, \textit{countZeros\_DC\_EQ\_m3}) which has a low ground truth similarity of 0.1842. Figure \ref{fig:Fig 9} shows the control flow graphs of calVariance and countZeros\_DC\_EQ\_m3 functions respectively. The proposed funcGNN model predicted the similarity value for this pair as 0.1036 with an error of 0.0806

\begin{table*}
\centering
\setlength{\tabcolsep}{8pt} 
\renewcommand{\arraystretch}{1.2} 
\begin{tabular}{|p{1cm}|c|c|c|c|c|} 

 \hline
 \textbf{Case Study} & \textbf{Program 1} & \textbf{Program 2} & \textbf{Ground Truth} &  \textbf{Prediction} & \textbf{Error} \\
 \hline
 1 & elementwiseMax\_DC\_EQ\_m3 & elementwiseMax\_DC\_EQ\_m3 & 1.0 & 0.9732 & 0.0268\\
 \hline
 2 & heapSort\_L\_EQ\_m4 & bitwiseOr\_L\_EQ\_m4 & 0.0108& 0.0035 & 0.0073\\
 \hline
 3 & calVariance & countZeros\_DC\_EQ\_m3 & 0.1842 & 0.1036 & 0.0806\\
 \hline
\end{tabular}
\caption{Case study observations from the test dataset}
\label{table:3}
\end{table*}
\section{Limitations and Threats to Validity}

There exist multiple factors that could be considered threats to the validity of our results. These include: 
\begin{itemize}
    \item \textbf{CFG creation} In this study we have used the open source tool Soot for generating the CFGs. Though the Jimple address format provided by Soot gives a detailed atomic level representation of a program, it can only analyse Java programs. Thus the scope of this project is limited to Java programs and repositories. 
    \item \textbf{Data Variability} We had initially taken a small set of Java programs and mutated them to create four sets of similar variants. The reason for this approach was to have examples of programs with small GED values in the dataset. This comes with the drawback that it reduces the variability in the structure of programs in the dataset. However, the idea of this study was to understand the approximate GED among program graph pairs and not its logic or working. 
    \item \textbf{Program Size } Since the GED problem is NP-hard, we used individual Java unit functions as our dataset and not the entire Java class file, in order to reduce the number of nodes in each graph. It would be interesting to see how our approach generalises to predicting the approximate GED of large Java class files. 
    \item \textbf{Loss by Approximation} We have employed an approximate value of the actual GED by converting it to a Quadratic Assignment Problem (QAP)\cite{bougleux2017graph}. Though this provides an estimate close to the actual GED score, there is always a trade-off when we prioritise computation time over accuracy. Thus our approach possess all the limitations of the GED calculation used in the QAP approximation method. 
    \item \textbf{Backpropagation of Histogram} In our \textit{Bottom-up} approach we extracted the histogram feature representation by performing atomic level node comparison. However, histograms cannot be trained using the backpropagation methodology as there is not a continuous differential function. We use the histogram features just to enhance the global graph features as in \cite{bai2019simgnn}. 

\end{itemize}
\section{Related Work}

The task of program or code similarity is of fundamental interest in software engineering, and can be traced back to \cite{berghel1984measurements}. Program similarity is applied in many software engineering problems, including code plagiarism\cite{faidhi1987empirical,zhang2014program}, authorship identification \cite{dasgupta2010not, kalgutkar2019code}, code search \cite{niu2017learning, keivanloo2014spotting}, clone identification and refactoring \cite{krishnan2014unification, zibran2013conflict},  and detecting malware patterns \cite{karnik2007detecting, cesare2011malware}. \cite{walenstein2007similarity} provides a detailed review of numerous methodologies to estimate program similarity and compare code. 

There has been some effort to solve the problem of program similarity using traditional machine learning techniques. \cite{maletic2000using, kim2015machine, hoste2006performance, phansalkar2005measuring}. However all these studies used features hand-crafted manually by domain experts, which is expensive in terms of time and expertise. The use of deep learning techniques which avoid this feature engineering approach, was applied for solving program similarity in \cite{marastoni2018deep}. One drawback of deep learning is the amount of tagged data it requires for training, which might be difficult to obtain in the field of software engineering. \cite{nair2019leveraging} demonstrates the use of equivalent mutants as a data augmentation method on source code, to alleviate the data crunch problem. In our approach we have used this data augmentation methodology for creating program pairs with low GED score.  

Another limitation of deep learning models is that, they are mainly trained on the shallow textual structure of a program (syntax), and can miss out on semantic features \cite{allamanis2017learning}. The study in \cite{allamanis2017learning} suggests that using graphs we can represent both the syntactic and semantic structure of code, and it demonstrates the effectiveness of graph neural networks for better program analysis. 

Graphs, especially control flow graphs, have been used extensively to solve many software engineering problems \cite{krinke2006mining, vujovsevic2013software, feng2016scalable,kanewala2016predicting, nandi2016anomaly, phan2017convolutional}. In \cite{zhao2018deepsim}, labeled CFGs were analysed using deep learning techniques for learning code semantic similarity. However, the input to the deep learning model in this study was a hand-crafted feature matrix thus restricting the models's capability of inferring the semantics of the graph.  Graph neural networks (GNNs) have emerged as a successful class of neural network models capable of learning of graph representations effectively \cite{gori2005new, kipf2016semi, scarselli2008graph, hamilton2017inductive, xu2018powerful}. In our approach we have harnessed the graph learning capability of GNNs to analyse labeled CFG graphs for program similarity. 

The most similar work to our approach is \cite{bai2019simgnn} where the authors propose SimGNN for graph similarity. The main differences between these two studies are: (i) the neighbourhood aggregation method (GCN\cite{kipf2016semi} in SimGNN and GraphSAGE\cite{hamilton2017inductive} in ours), (ii) the choice of hyperparameters and the activation functions used, and (iii) the dataset used. Regarding (iii), in \cite{bai2019simgnn} the authors evaluate their model on unlabeled program dependency graphs (PDG) of C programs. All the unlabeled nodes in their approach were initialised with the same label, leading to each node having the same initial embedding representation. Hence the only code features learned were the data flow features and not the node features (as in our approach). 
In our approach we trained on labeled CFGs where each node in the CFG was labeled with an atomic program statement. Each such atomic operation was initialised with an unique embedding, thus providing the learned model with richer program structure. Also the dataset of \cite{bai2019simgnn} consisted of small programs each restricted to a maximum value of 10. In our approach, the model was trained on larger program graphs with an average size of 19.76 and a maximum size of 72 (see Figure \ref{fig:Fig 4}).  

Other interesting contributions similar to ours are \cite{ xu2017neural, li2019graph}, where labeled CFGs are analysed to finding binary code similarities and to check the presence of any known security vulnerabilities in it. However, here each node in the CFG dataset represents multiple attributes (such as mov, lea, cmp and jbe all in one node). This makes it difficult to calculate an appropriate embedding to that node which summarizes all of its operations. In our approach, each atomic operation in the program is assigned a new node to achieve a better node embedding.   
 
\section{Conclusion}

In this paper, we have studied the problem of program similarity using graph edit distance and proposed \textit{funcGNN} a graph neural network approach to estimating the GED. To characterise the semantics and logic of the program we analyse its control flow graph representation. \textit{funcGNN} inherits the inductive and node-order invariant properties of graph convolutional networks and uses it to create a semantically rich embedding for each node in a CFG. The evaluation study carried out in our research shows that \textit{funcGNN} is capable of estimating the approximate graph edit distance of unseen program pairs with very low error rate and is computationally efficient. We have discussed the limitations and drawbacks of our approach, and we see potential improvements in future. We will also consider how to apply \textit{funcGNN} for solving related software engineering challenges such as clone refactoring, for which finding program similarity is crucial. 

\section{Acknowledgments}
 The authors gratefully acknowledge financial support from the ITEA3 TESTOMAT Project 16032 and Ericsson AB.

\bibliographystyle{ACM-Reference-Format}
\bibliography{sample-base}


\end{document}